\documentclass[runningheads]{llncs}
\usepackage{graphicx}
\usepackage{subfigure}
\usepackage{amsmath}
\usepackage{makeidx}
\begin{document}
\title{Real-time Surface Deformation Recovery\\from Stereo Videos}
%
%
\author{Haoyin Zhou\and
Jayender Jagadeesan}
\authorrunning{H. Zhou et al.}
%
\institute{Surgical Planning Laboratory,\\Brigham and Women's Hospital, Harvard Medical School, Boston, MA, 02115, USA}
\maketitle              
\begin{abstract}
Tissue deformation during the surgery may significantly decrease the accuracy of surgical navigation systems. In this paper, we propose an approach to estimate the deformation of tissue surface from stereo videos in real-time, which is capable of handling occlusion, smooth surface and fast deformation. We first use a stereo matching method to extract depth information from stereo video frames and generate the tissue template, and then estimate the deformation of the obtained template by minimizing ICP, ORB feature matching and as-rigid-as-possible (ARAP) costs. The main novelties are twofold: (1) Due to non-rigid deformation, feature matching outliers are difficult to be removed by traditional RANSAC methods; therefore we propose a novel 1-point RANSAC and reweighting method to preselect matching inliers, which handles smooth surfaces and fast deformations. (2) We propose a novel ARAP cost function based on dense connections between the control points to achieve better smoothing performance with limited number of iterations. Algorithms are designed and implemented for GPU parallel computing. Experiments on \textit{ex-} and \textit{in vivo} data showed that this approach works at an update rate of 15Hz with an accuracy of less than 2.5 mm on a NVIDIA Titan X GPU.

\end{abstract}
\section{Background}

Tissue visualization during surgery is typically limited to the anatomical surface exposed to the surgeon through an optical imaging modality, such as laparoscope, endoscope or microscope. To identify the critical structures lying below the tissue surface, surgical navigation systems need to register the intraoperative data to preoperative MR/CT imaging before surgical resection. However, during surgery, tissue deformation caused by heartbeat, respiration and instruments interaction may make the initial registration results less accurate. The ability to compensate for tissue deformation is essential for improving the accuracy of surgical navigation. In this paper, we propose an approach to recover the deformation of tissue surface from stereo optical videos in real-time.

In recent years, several groups have investigated methods to recover tissue deformation from optical videos, and most methods are based on the minimization of non-rigid matching and smoothing costs \cite{SchoobMedImgAnalysis2017}. For example, Collins et al proposed a monocular vision-based method that first generated the tissue template and then estimated the template deformation by matching the texture and boundaries with a non-rigid iterative closet points (ICP) method \cite{CollinsMiccai2016}. In this method, the non-rigid ICP-based boundary matching algorithm significantly improves the accuracy. However, during surgery, only a small area of the target tissue may be exposed and the boundaries are often invisible, which makes it difficult to match the template. Object deformation recovery in the computer vision field is also a suitable approach to recover tissue deformation. For example, Zollhöfer et al proposed to generate the template from an RGB-D camera and then track the deformation by minimizing non-rigid ICP, color and smoothing costs \cite{MicrosoftNonRigid2014}. Newcombe et al have developed a novel deformation recovery method that does not require the initial template and uses sparse control points to represent the deformation \cite{Newcombe}. Guo et al used forward and backward $L_0$ regularization to refine the deformation recovery results \cite{GuoICCV2015}. To date, most deformation recovery methods \cite{ModrzejewskiMICCAI2018}\cite{PetitIROS2015} are based on the non-rigid ICP alignment to obtain matching information between the template and the current input, such as monocular/stereo videos or 3D point clouds from RGB-D sensors. However, non-rigid ICP suffers from a drawback that it cannot track fast tissue deformation and camera motion, and obtain accurate alignment in the tangential directions on smooth tissue surfaces. During surgery, the endo/laparoscope may move fast or even temporally out of the patient for cleaning, which makes non-rigid ICP difficult to track the tissue. In addition, smoke and blood during the surgery may cause significant occlusion and interfere with the tracking process. Hence, the ability to match the template and the input video when non-rigid deformation exists is essential for intraoperative use of deformation recovery methods.


A natural idea to obtain additional information is to match the feature points between the template and the input video. Among many types of feature descriptors, ORB \cite{ORB} has been widely used in real-time applications due to its efficiency. To handle feature matching outliers, RANSAC-based methods have proven effective in rigid scenarios but are difficult to handle non-rigid deformation \cite{DefendNonRigidRANSAC}. Another common method to address outliers is to apply robust kernels to the cost function, which cannot handle fast motion. In this paper, we propose a novel method that combines 1-point-RANSAC and reweighting methods to handle matching outliers in non-rigid scenarios. In addition, we propose a novel as-rigid-as-possible (ARAP) \cite{ARAP2007} method based on dense connections to achieve better smoothing performance with limited number of iterations.

\section{Method}

\begin{figure} [h]
\vspace{0.0cm}
\centering
 \includegraphics[width=.85\textwidth]{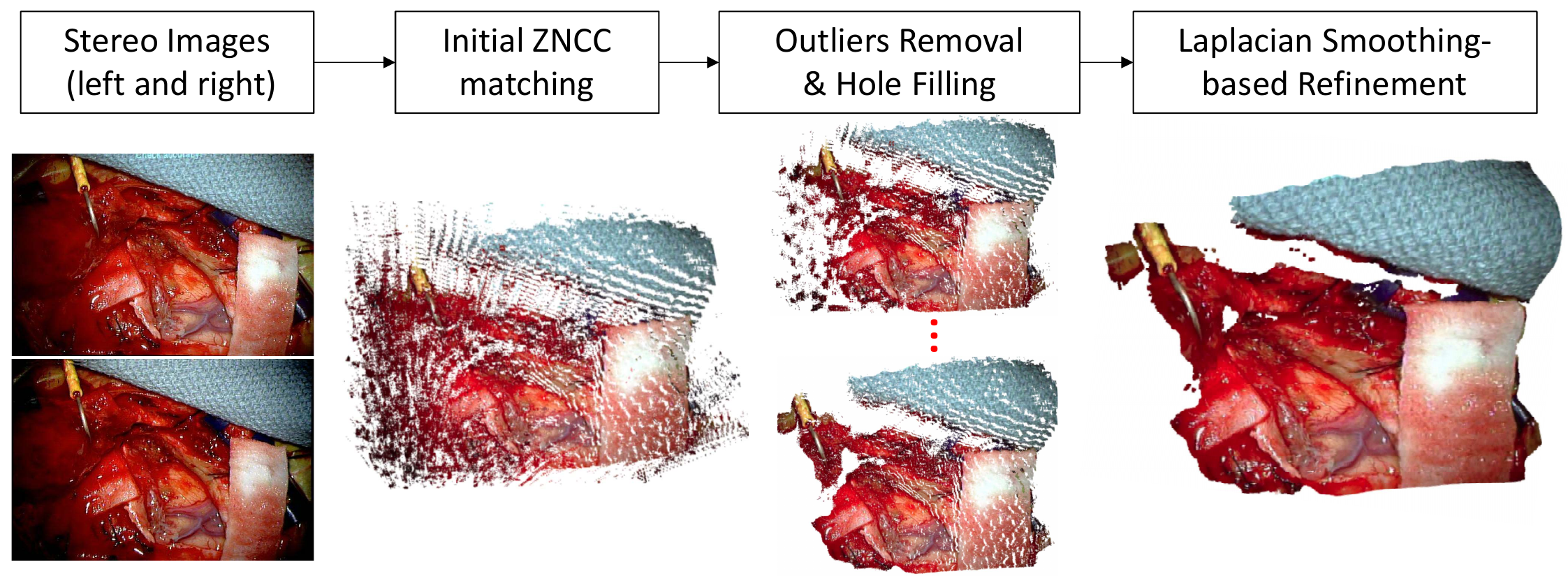}
 \caption{The process of our stereo matching method with a pair of stereo microscopy images captured during neurosurgery.}
\label{fig_BinocularProcess}
\end{figure}

As shown in Fig. \ref{fig_BinocularProcess}, we proposed a GPU-based stereo matching method, which includes several efficient post-processing steps to extract 3D information from stereo videos in real-time. Readers may refer to Ref. \cite{ZHOUTMI2019} for more details on this stereo matching method.

In our system, the initial template of the tissue surface is generated by the stereo matching method, then we track the deformation of the template by representing the non-rigid deformation with sparse control points on the template, and estimating the parameters of the control points to make the deformed template match the output of the GPU-based stereo matching method. The algorithms are parallelized and run on the GPU. Similar to DynamicFusion \cite{Newcombe}, we employ dual-quaternion to represent deformation and each control point $i$ is assigned a dual-quaternion $W_i^t$ to represent its warp function at time $t$, and the template points are deformed according to the interpolation of neighboring control points. Then, the deformation recovery problem is to estimate $W_i^t$, $i=1,...,N$, and we use the Levenberg-Marquardt algorithm to minimize the following cost function

\begin{equation}
f_{{\rm{Total}}}(W_i^t) = {f_{{\rm{ICP}}}} + {w_{{\rm{ORB}}}}{f_{{\rm{ORB}}}} + {w_{{\rm{ARAP}}}}{f_{{\rm{ARAP}}}},
\label{eq_totalcost}
\end{equation}

\noindent where ${f_{{\rm{ICP}}}}$ and ${f_{{\rm{ORB}}}}$ are based on non-rigid ICP and ORB matches between the template and the current stereo matching results respectively. The as-rigid-as-possible (ARAP) cost ${f_{{\rm{ARAP}}}}$ smoothes the estimated warp functions $W_i^t$, which is especially important for the estimation of occluded areas. ${w_{{\rm{ORB}}}}$ and ${w_{{\rm{ARAP}}}}$ are user defined weights. In our experiments, we use ${w_{{\rm{ORB}}}} = 10.0$ and ${w_{{\rm{ARAP}}}}$ is dynamically adjusted due to the varying number of valid points in $f_{\rm{ICP}}$ and ORB matching inliers in ${f_{{\rm{ORB}}}}$. We sum up the related weights of ICP and ORB terms for each $W_i^t$, and scale up or down $w_{\rm{ARAP}}$ accordingly.

A GPU-based parallel Levenberg-Marquardt (LM) algorithm was developed to minimize the cost \eqref{eq_totalcost}. We update each $W_i^t$ independently in the LM iterations. For the computation of the Jacobian matrix $\bf{J}$ related to each $W_i^t$, multiple parallel GPU threads are launched to compute rows of $\bf{J}$, then we perform Cholesky decomposition to update $W_i^t$, $i=1,...,N$.

The non-rigid ICP term ${f_{{\rm{ICP}}}}$ is determined by the distances between the deformed template and the stereo matching results. The Tukey's penalty function is employed to handle outliers. We have developed a rasterization process that re-projects the template points to the imaging plane to build correspondences between template points and the stereo matching results, which is parallelized to each template point and runs on the GPU. This rasterization step is faster than kd-tree-based closest points search in the 3D space. Only the distance component in the normal directions are considered, which avoids the problem that non-rigid ICP is inaccurate in the tangential directions when aligning smooth surfaces.

\subsection{ORB Feature Matching and Inliers Pre-selection}

\begin{figure} [h]
\vspace{0.0cm}
\centering
  \includegraphics[width=0.99\textwidth]{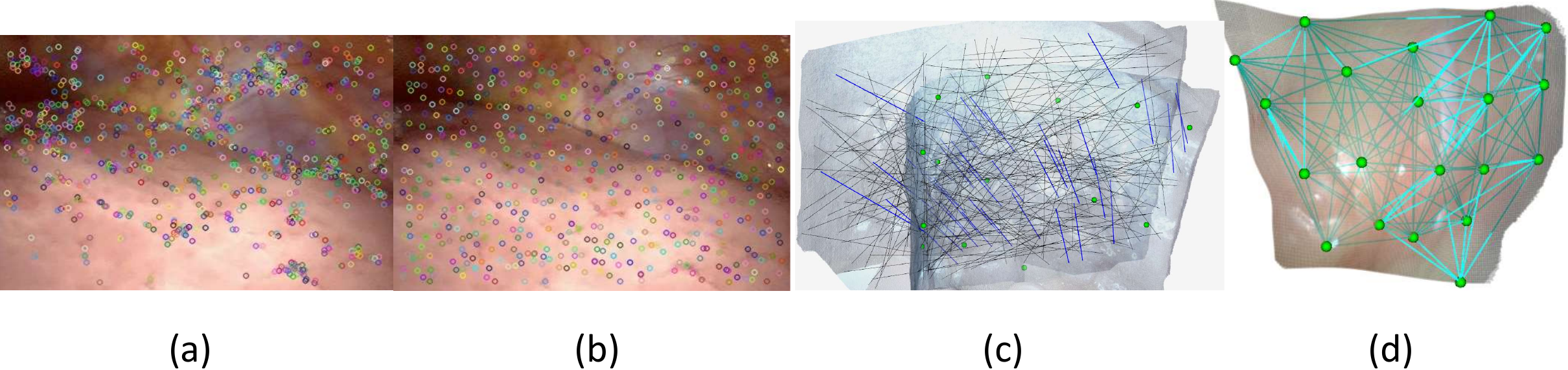}
  \caption{(a)-(b) ORB feature detection results on laparoscopy images captured during a lung surgery using (a) OpenCV (b) Our method. (c) Matching inliers pre-selection results with a deforming phantom. The blue lines are selected inliers and black lines are identified as outliers. (d) Dense connections between control points with a silicon heart phantom.}
\label{fig_Feature}
\end{figure}

As shown in Fig. \ref{fig_Feature}(a)-(b), standard ORB feature detection concentrates on rich texture areas, which may lead to the lack of matching information at low texture areas. Hence, we first develop a method to detect uniform ORB features to improve the accuracy of deformation recovery, which uses GPU to detect FAST corners and suppresses those if a neighboring pixel has larger corner response in parallel. Then, the ORB features of the initial template are matched to the live video frames. Two corresponding 3D point clouds are obtained, which may include incorrect matches.

Since at least three matches are needed to determine the rigid relative pose between two 3D point clouds, traditional RANSAC methods only work when the three matches are all inliers and have similar deformation \cite{DefendNonRigidRANSAC}. Another common method to handle outliers is to apply robust kernels to the cost function, which is effective but cannot handle fast camera motion or tissue deformation. Under a reasonable assumption that local deformations at small areas of the tissue surface are approximate to rigid transforms, we propose a novel 1-point-RANSAC and reweighting method to pre-select potential matching inliers following the idea of Ref. \cite{R1PPnP}, as shown in Fig. \ref{fig_Feature}(c). Denoting the two sets of corresponding 3D ORB features as $o_k^1$ and $o_k^2$, $k = 1,...,N$, a random match $k_0$ is selected as the reference, and rectify the coordinates with respect to $k_0$ by

\begin{equation}
{\bf{S}}_{k0}^l = {\left[ {\begin{array}{*{20}{c}}
{o_1^l - o_{k0}^l,}& \cdots &{,o_N^l - o_{k0}^l}
\end{array}} \right]_{3 \times N}},l = 1,2.
\end{equation}

For a reference $k_0$, we denote the local rigid transform as $o_{k0}^2 = {\bf{R}}o_{k0}^1 + {\bf{T}}$, where ${\bf{R}} \in SO(3)$ is the rotation matrix and ${\bf{T}}$ is the translation vector. Rigid transform for a neighboring match inlier $k$ should satisfy

\begin{equation}
 {\bf{S}}_{k0}^2(k) \approx {\bf{RS}}_{k0}^1(k),
\label{eq_sk}
\end{equation}

\noindent where ${\bf{S}}_{k0}(k)$ is the $k$th column of ${\bf{S}}_{k0}$, and $\bf{R}$ can be obtained from matches that satisfy \eqref{eq_sk}. We propose a reweighting method to eliminate the impacts of other matches, that is

\begin{equation}
{d_k} = \left\| {{{\bf{S}}_{k0}^2}(k) - {\bf{R}}{{\bf{S}}_{k0}^1}(k)} \right\|, {w_k} = \min \left( {H/{d_k},1} \right),
\label{eq_updatewk}
\end{equation}

\noindent where $d_k$ is the distance related to the $k$th match. $w_k$ is the weight of the $k$th ORB match and if the $k$th match is either an outlier, or an inlier that does not satisfy \eqref{eq_sk}, $w_k$ is small. $H$ is a predefined threshold. With a selected reference $k_0$, we alternatively update $\bf{R}$ from weighted ${\bf{S}}_{k0}^1$ and ${\bf{S}}_{k0}^2$, and update $w_k$ according to \eqref{eq_updatewk}. In experiments we perform 10 iterations with each $k_0$. A small sum of $w_k$ suggests that few matches satisfy \eqref{eq_sk} and $k_0$ may be an outlier, and we omit the results with reference $k_0$. In our experiments, we randomly select 30 different matches as the reference $k_0$.

We first apply this 1-point-RANSAC + reweighting method to assign weights to ORB matches, the results of which will be used in the subsequent LM algorithm to minimize term ${f_{{\rm{ORB}}}}$ in \eqref{eq_totalcost}. It should be clarified that we are not implying that this 1-point-RANSAC + reweighting method is able to find all inliers. To take into account all inliers, in the LM algorithm we assign the pre-selected matches the same weight as $w_k$, and assign other ORB matches weight according to ${w_k} =  - 1/(5H){d_k} + 1$, ${w_k} \in \left[ {0,1} \right]$.

\subsection{As-Rigid-As-Possible Smoothing}

Traditional ARAP methods are based on sparse connections, such as triangular meshes. This type of connection is too sparse to propagate the smoothing impact fast enough, and in practice we found that it cannot perform well with the limited number of iterations in the LM algorithm. Hence, we propose to use dense connections as shown in Fig. \ref{fig_Feature}(d). The weights of connections in traditional ARAP methods are sensitive and need to be specifically designed based on the angles of the triangular mesh \cite{ARAP2007}, hence the ARAP cost function has to be redesigned to handle the dense connections as follows:

\begin{equation}
{f_{{\rm{ARAP}}}} = \sum\limits_{i1,i2} {{w_{i1,i2}}\left( {f_{{\rm{length,}}i1i2}} + {{w_{{\rm{angle}}}}{f_{{\rm{angle,}}i1i2}} + {w_{{\rm{rotation}}}}{f_{{\rm{rotation,}}i1i2}}} \right)}
\end{equation}

\noindent where $i_1$ and $i_2$ are two control points. $w_{i1,i2}$ is the weight of connection between $i_1$ and $i_2$, and a smaller distance between points $i_1$ and $i_2$ at time 0 suggests larger $w_{i1,i2}$. We use ${w_{{\rm{angle}}}} = 20.0$ and ${w_{{\rm{rotation}}}} = 100.0$.

For control points $i_1$ and $i_2$,

\begin{equation}
\begin{array}{l}
{f_{{\rm{length,}}i1i2}} = {\left( {\left\| {p_{i2}^t - p_{i1}^t} \right\| - \left\| {p_{i2}^0 - p_{i1}^0} \right\|} \right)^2}\\
{f_{{\rm{angle,}}i1i2}} = {\mathop{\rm acos}\nolimits} (W_{i1}^t(p_{i2}^0) - p_{i1}^t,p_{i2}^t - p_{i1}^t)\\
{f_{{\rm{rotation,}}i1i2}} = {\left\| {W_{i1}^t(1,2,3,4) - W_{i2}^t(1,2,3,4)} \right\|^2}
\end{array}
\end{equation}

\noindent where $p_i^t$ is the coordinate of point $i$ at time $t$. ${f_{{\rm{angle,}}i1i2}}$ equals to the angle between the normalized vectors $W_{i1}^t(p_{i2}^0) - p_{i1}^t$ and $p_{i2}^t - p_{i1}^t$, where $W_{i1}^t(p_{i2}^0)$ suggest to apply $W_{i1}^t$ to $p_{i2}^0$. ${f_{{\rm{rotation,}}i1i2}}$ is introduced because $W_i^t$ has 6-DoFs, which is determined by the differences between the first four components of dual-quaternion $W_{i1}^t$ and $W_{i2}^t$.

\section{Experiments}

\begin{figure} [h]
\vspace{0.0cm}
\centering
  \includegraphics[width=0.90\textwidth] {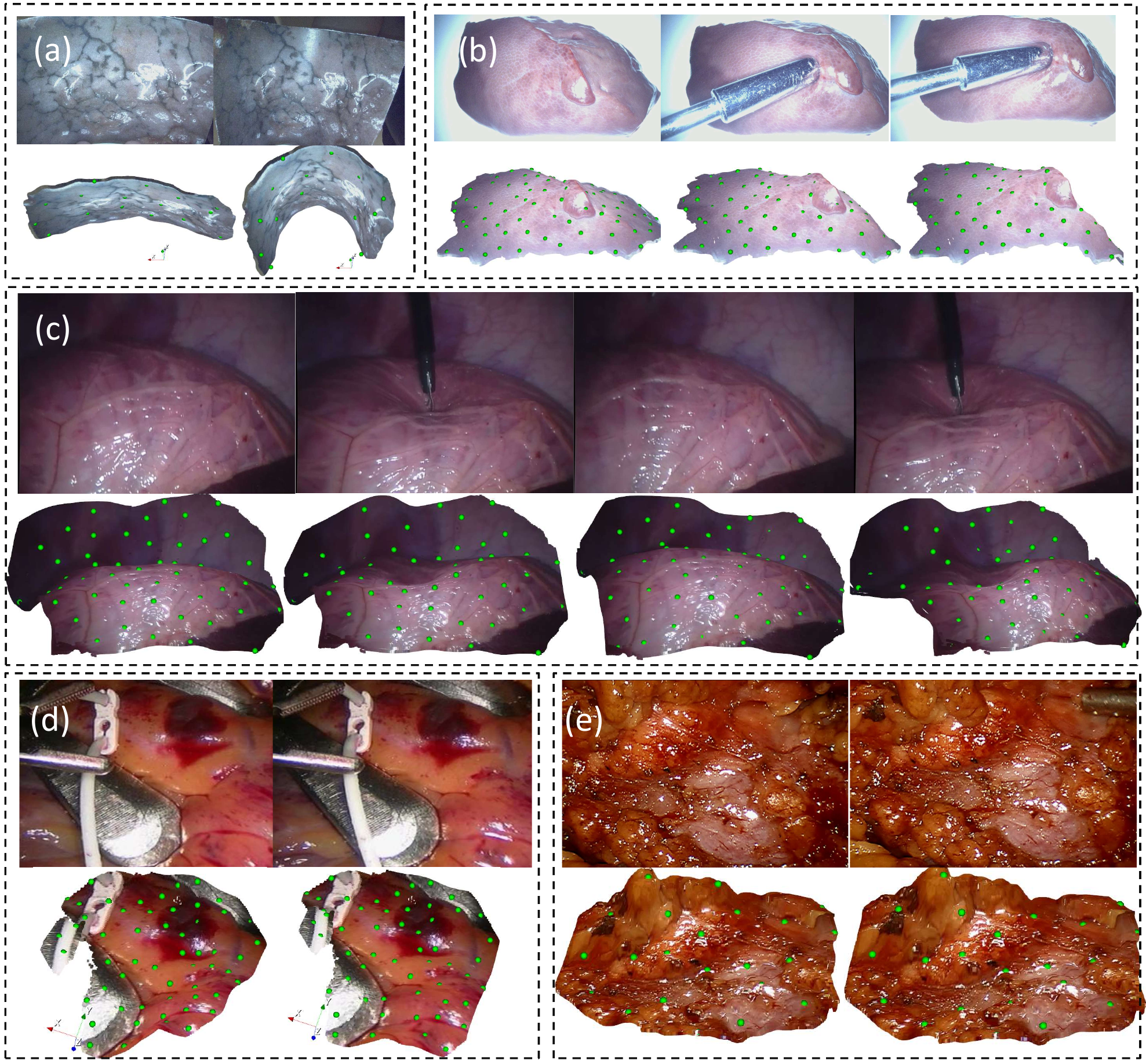} 
  \caption{Qualitative experiments. First row: input video frames. Second row: the deformed template and the control points (green dots). (a) Phantom. (b) \textit{Ex vivo} porcine liver. (c) Hamlyn \textit{in vivo} data with deformation caused by instrument interaction (d) Hamlyn \textit{in vivo} data with respiration and heartbeat. (e) \textit{In vivo} kidney data with deformation caused by respiration.}
\label{fig_qualification}
\end{figure}

\begin{figure} [h]
\vspace{0.0cm}
\centering
  \includegraphics[width=1.0\textwidth]{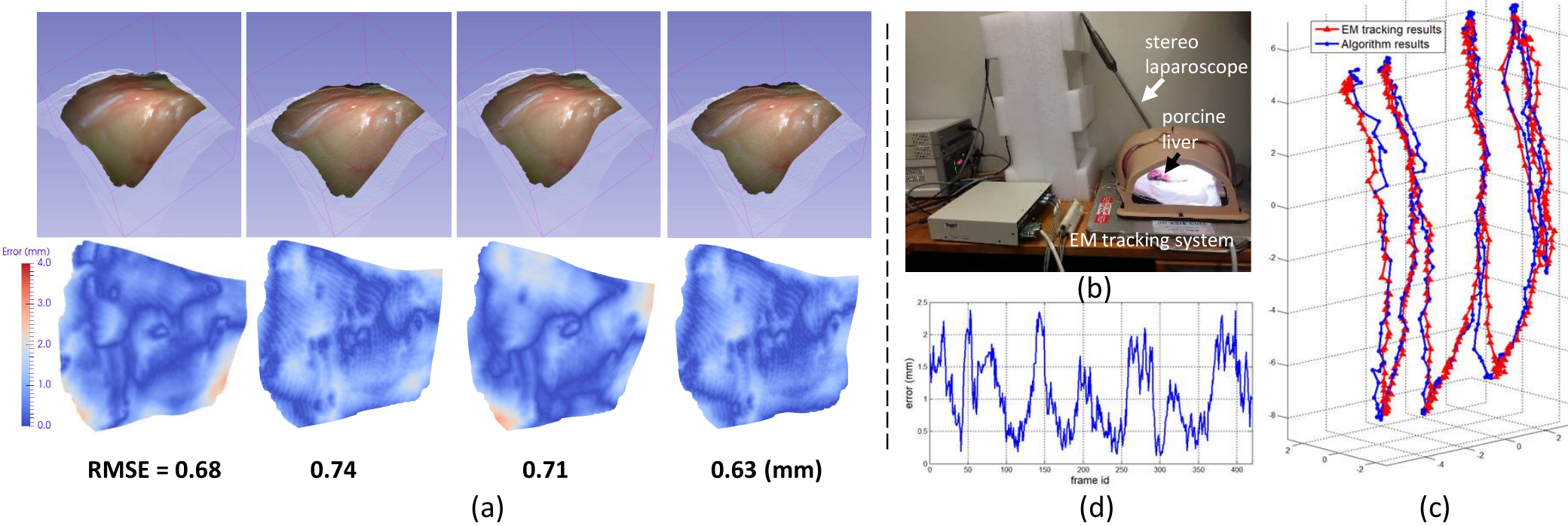}
  \caption{Quantitative experiments. (a) Hamlyn heart Phantom data. First row: colored models are the deformed templates, white points are the ground truth. Second row: distance maps. Average runtime: stereo matching 3.8 ms, ORB feature detection and matching 10.6 ms, inliers pre-selection 4.1 ms, LM 14.2 ms. (b)-(d) Experiment with the EM tracking system. (b) Hardware. (c) 3D trajectories. (d) Errors. Average runtime: stereo matching 17.6 ms, ORB feature detection and matching 11.6 ms, inliers pre-selection 3.1 ms, LM 30.7 ms.}
\label{fig_quantitative}
\end{figure}

Algorithms were implemented with CUDA C++ running on a desktop with Intel Xeon 3.0 GHz CPU and NVIDIA Titan X GPU. We first conducted qualitative experiments on \textit{ex-} and \textit{in vivo} data. As shown in Fig \ref{fig_qualification}(a), we deformed a smooth phantom with lung surface texture and captured $960 \times 540$ stereo videos with a KARL STORZ stereo laparoscope. We removed intermediate video frames between the two frames in Fig \ref{fig_qualification}(a) to simulate fast deformation, and our method is capable of tracking the large deformation. The second experiment was conducted with \textit{ex vivo} porcine liver as shown in Fig. \ref{fig_qualification}(b). The deformation was caused by instrument interaction, and our method is able to handle instrument occlusion. For the \textit{in vivo} experiments shown in Fig. \ref{fig_qualification}(c)-(e), we used both the Hamlyn data \cite{HamlynPaper} and our data, in which the videos have camera motion and tissue deformation. We generated the tissue template before instrument interaction and then track the deformation of the template. The algorithm detected key inlier ORB features on the reconstructed surface and tracked these template features robustly in spite of respiratory and pulsatile motions, and instrument occlusions. These results highlight the robustness of tracking in spite of physiological motions and varying illumination.

We conducted two quantitative experiments. The first experiment was conducted on Hamlyn data as shown in Fig. \ref{fig_quantitative}(a). The Hamlyn data consists of stereo video images of a silicon phantom simulating heartbeat motion and corresponding ground truth was obtained using CT scan. The template was generated from the first video frame. Results show an RMSE of less than 1mm and the average runtime of 32.7ms per frame. In the second experiment, we used the EM tracking system (medSAFE Ascension Technologies Inc.) as the ground truth, as shown in Fig. \ref{fig_quantitative}(b)-(d). The porcine liver was placed in an abdominal phantom (The Chamberlain Group) and a medSAFE EM sensor was attached to the liver surface. We deformed the liver manually and recorded the EM sensor measurements and compared it with that of the our method. Deformation estimation results on 420 video frames (Fig. \ref{fig_quantitative}(c)-(d)) show a mean error of 1.06mm and standard deviation of 0.56mm. As shown in Fig. \ref{fig_quantitative}(c), the maximum distance between the trajectory points is 15.7mm. The average runtime was 63.0ms per frame.

\section{Conclusion}

We propose a novel deformation recovery method that integrates the ORB feature, which is able to handle fast motion, smooth surfaces and occlusion. The limitation of this work is that it strongly relies on ORB feature matching, which may fail when the deformation is extremely large and different light reflection may make it difficult to obtain enough number of ORB matching inliers.


%

%
\section{Acknowledgments}

This project was supported by the National Institute of Biomedical Imaging and Bioengineering of the National Institutes of Health through Grant Numbers P41EB015898 and R01EB025964. Unrelated to this publication, Jayender Jagadeesan owns equity in Navigation Sciences,Inc. He is a co-inventor of a navigation device to assist surgeons in tumor excision that is licensed to Navigation Sciences. Dr. Jagadeesans interests were reviewed and are managed by BWH and Partners HealthCare in accordance with their conflict of interest policies.

%
%
%

\end{document}